\documentclass[11pt]{article}
\usepackage[T1]{fontenc}
\usepackage{amsthm, amsmath, amssymb, amsfonts, url, booktabs, tikz, setspace, fancyhdr, bm, mathrsfs}
\usepackage{geometry}
\usepackage{hyperref, enumerate}
\usepackage[shortlabels]{enumitem}
\usepackage[babel]{microtype}
\usepackage[english]{babel}
\usepackage[capitalise]{cleveref}
\usepackage{comment}
\usepackage{bbm}

\usepackage{booktabs}
\usepackage{csquotes}
\usepackage{mathabx}
\usepackage{graphicx}
\usepackage{float}
\usepackage{subcaption}

\counterwithin{figure}{section}

\theoremstyle{definition}

\newtheorem*{defn-non}{Definition}

\newlist{Case}{enumerate}{3}
\setlist[Case, 1]{%
    label           =   {\bfseries Case \arabic*.},
    labelindent=1em ,labelwidth=1cm, labelsep*=1em, leftmargin =!
}
\setlist[Case, 2]{%
    label           =   {\bfseries Subcase \arabic{Casei}.\arabic*.},
    labelindent=-1em ,labelwidth=1cm, labelsep*=1em, leftmargin =!
}
\setlist[Case, 3]{%
    label           =   {\bfseries Subsubcase \arabic{Casei}.\arabic{Caseii}.\arabic*.},
    labelindent=-1em ,labelwidth=1cm, labelsep*=1em, leftmargin =!
}

\usepackage{todonotes} 
\usepackage{cite}

\title{Deep Learning for On-Street Parking Violation Prediction}

\author{
Vo Thien Nhan \thanks{
Institute of Engineering, Ho Chi Minh City University of Technology (HUTECH), Vietnam \\  Email: thiennhan.math@gmail.com}}

\begin{document}
\maketitle
\begin{abstract}
Illegal parking along with the lack of available parking spaces are among the biggest issues faced in many large cities. These issues can have a significant impact on the quality of life of citizens. On-street parking systems have been designed to this end aiming at ensuring that parking spaces will be available for the local population, while also providing easy access to parking for people visiting the city center. However, these systems are often affected by illegal parking, providing incorrect information regarding the availability of parking spaces. Even though this can be mitigated using sensors for detecting the presence of cars in various parking sectors, the cost of these implementations is usually prohibiting large. In this paper, we investigate an indirect way of predicting parking violations at a fine-grained level, equipping such parking systems with a valuable tool for providing more accurate information to citizens. To this end, we employed a Deep Learning (DL)-based model to predict fine-grained parking violation rates for on-street parking systems. Moreover, we developed a data augmentation and smoothing technique for further improving the accuracy of DL models under the presence of missing and noisy data. We demonstrate, using experiments on real data collected in Thessaloniki, Greece, that the developed system can indeed provide accurate parking violation predictions.
\end{abstract}
\section{Introduction}
Searching for a parking space in a large city is often very challenging and frustrating for drivers, especially in crowded areas like the city’s centre or around of points of interest. To address this issue, many cities have implemented intelligent on-street parking systems \cite{ref1}. In such systems, available park- ing spaces are often divided into sectors, with each sector being composed of several parking slots. Drivers usually manually update the system by stating the slot/sector where they have parked, when paying the parking fee. In this way, the parking system is updated and can provide useful information to drivers in order to guide them to sectors that have free slots available. However, when drivers park without providing the necessary information to the system, i.e., usually to avoid paying the parking fee, the system no longer has up-to-date information regarding the availability of parking slots and cannot provide useful information to other drivers. During peak hours, such phenomena can often lead to a significant degradation in the system’s performance, which lowers the trust of drivers to such systems and discourages their use.

The aforementioned issue can be addressed by installing
street occupation sensors that can detect the presence of cars
in various parking sectors, e.g., either street sensors buried
under the road or vision-based systems \cite{ref2}. However, the cost of installing and maintaining such complicated systems
in large cities can outweigh the benefits of their installation. Furthermore, such systems tend to work better in more
constrained settings, e.g., in closed spaces, instead of public
parking systems that can span an entire city. In this work, we
investigate whether it is possible to acquire such information
using indirect measurements, such as weather information,
time and date, as well as historical data. More specifically, we
aim to predict illegal parking rates in different parking sectors
around a city. This information, along with the existing information held by such systems (i.e., cars that are legally parked)
would allow for upgrading their operation with minimal cost.
Indeed, such an indirect prediction system would allow for
providing corrections to the number of available parking slots,
taking into account slots that might have been taken up by
illegal parking, redirecting users towards sectors that are more
probable to have free parking slots

To this end, in this work we aim on developing a ma- chine learning-based approach for predicting illegal parking in various sectors of an on-street parking system. However, de- veloping such a system poses significant challenges regarding several aspects of its design, ranging from how the input data are gathered and encoded to how the ground truth parking violation data are collected. Despite the importance of this problem, there is a relatively small number of works attempting to address this issue. For example, in \cite{ref3} parking violations are predicted using random forests and existing open data, while \cite{ref4} tackles the problem of predicting double parking events. However, the latter only focuses on double parking events, while the former, despite solving a similar problem, provides coarse information. In contrast, this paper focuses on solving a fine-grained variant of the on-street parking violation prediction providing information for hundreds of sectors. Other approaches, such as \cite{ref5}, \cite{ref6}, attempt to directly predict the number of available parking spaces instead of focusing only on the illegal parking rate, potentially solving a significantly harder problem. It is worth noting that most of these approaches are often tailored to the specifics of each parking system, available annotation type, and city structure, so it is not always possible to directly apply them in other scenarios.

The main aim of this work is to develop and evaluate a Deep Learning (DL)-based on-street parking violation pre- diction system that works integrated with on-street parking systems. DL has been shown to outperform other approaches for a wide variety of tasks \cite{ref7}, yet its application to novel

\begin{figure}[H]
    \centering
    \includegraphics[width=0.5\linewidth]{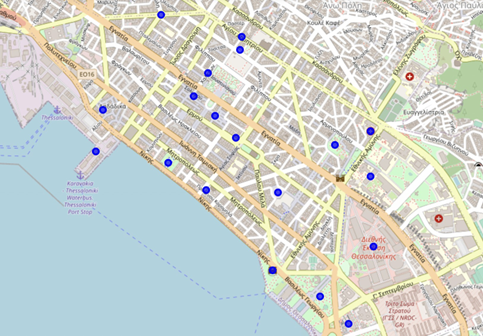}
    \caption{A total of 19 PoIs were defined and used for representing various sectors of the used on-street parking system}
    \label{fig:1}
\end{figure}
domains remains challenging. Among the main challenges that
we face is the existence of sparse annotations, i.e., information
regarding illegal parking is only available when the police
scan for such cases. This poses significant challenges since
annotations can be especially noisy, i.e., having information
for a sector but not for neighboring ones. Directly using such
information in DL-based systems is challenging, since it is
known that DL models can be affected by such types of
annotations, which often leads to applying label smoothing
and distillation methods to mitigate these effects \cite{ref8}, \cite{ref9}. To
address these issues, we developed a novel data smoothing
and augmentation approach that assigns existing ground truth
annotation to spatial sectors and time slots, leading to significant performance improvements. To the best of our knowledge,
this is the first time that such an approach is applied in the
context of an on-street parking violation prediction system.
Other challenges include imprinting input data, such as the
geographical location of each sector, time and other information, such as pandemic measures (i.e., due to COVID-19) in an
efficient manner. Finally, we extensively evaluate the proposed
method using data from the THESi system implemented in
Thessaloniki, Greece.

The rest of the paper is structured as follows. Section II
introduces the proposed method, detailing all important aspects
of data pre-processing and model construction. Then, the
experimental evaluation is provided in Section III. Finally,
conclusions are drawn in Section IV

\section{Proposed Method}
As described before, parking slots in public on-street
parking systems are divided into sectors. These sectors consist
the fundamental block for which we predict the rate of parking
violations. Given that the location and capacity of each sector
are known, we can directly use the parking violation predictions for each sector in order to guide drivers towards sectors
where parking slots will be most probably available. The first
challenge we faced was to represent the geographical place of
a sector in a way that would allow DL models to generalize. To
this end, we avoided directly using a global coordinate system
and we opted for indirectly encoding sector information using
distances from several Points of Interest (PoI). A PoI is often
visited by citizens on a large scale and, as a result, it is crowded most hours of the day. A PoI could be a sight, a central park,
a museum, the city hall, and others. Given that traffic and
demand for parking slots are higher across different PoIs, this
way of encoding the information can allow a DL model to
associate PoIs with expected parking violation rates, as well
as encode sectors using distances from these points. To this
end, we measured the distance between the center of each
sector and the selected PoIs and used this vector to represent
each sector. The way this information was encoded is shown in
Fig. 1, where blue circles represent the PoIs and the red circle
shows an example of a sector. A total of 19 points of interest
were employed, leading to a 19-dimensional representation for
each sector. The capacity of each sector was also included as
a feature.

Furthermore, we also included information regarding the
current weekday (Monday to Sunday), day (1 to 31), and
month (January to December) in order to allow the DL model
to encode historical information regarding the expected parking
violations. To capture the periodic nature of these features
and prevent discontinuities in the features we opted for using
sine-based encoding for all of these features. For example, the
weekday, denoted by $x_{\omega}$ was encoded as: 
\begin{equation}
x_w = \sin\left(2\pi \frac{w}{7}\right),
\end{equation}
where \( w \) denotes the current day using sequential integer encoding, i.e., 0 for Monday, 1 for Tuesday, etc. Similarly, we extracted a feature for the current day \( x_d \) as:
\begin{equation}
x_d = \sin\left(2\pi \frac{d - 1}{N_d}\right), \tag{2}
\end{equation}
where \( d \) is the current day (from 1 to 31) and \( N_d \) is the number of days for the current month. Moreover, the month feature \( x_m \) was calculated as:
\begin{equation}
x_m = \sin\left(2\pi \frac{m - 1}{12}\right), \tag{3}
\end{equation}
where \( m \) is the current month (ranging from 1 to 12). As we explain later in this section, the prediction horizon is one-hour. Therefore, each day was also divided into time slots (each with one-hour duration) and this information was also included in the extracted features. Note that time slots do not span the whole day, since parking control is enforced from 7:00 until 19:00. Therefore, we did not use sine encoding for these features and we directly imprinted the temporal location of each time slot as its distance from the time when parking control is enforced.

Apart from the location of each sector and time features, weather conditions also often affect traffic and demand for parking spaces. Therefore, we used the hourly values of temperature and humidity. Temperature is represented in Celsius degrees, while humidity is provided as a percentage over the maximum possible humidity. Given that weather features usually do not immediately affect the behavior of drivers, i.e., if the temperature rises, traffic is not immediately affected, we opted for encoding weather features as the average of windows. To this end, the current temperature feature \( x_T \) for a time step \( t \) was calculated as:
\begin{equation}
x_T = \frac{1}{W} \sum_{i=1}^{W} T_{t-i}, \tag{4}
\end{equation}
where \( T_t \) represents the temperature at the \( t \)-th time step and \( W \) denotes the window size used for the averaging. Note that we omitted the time step index \( t \) from the feature notation to reduce clutter. In this work, we used the current and 5 previous
temperature measurements, i.e., $W =6$. Humidity features $x_h$ 
were similarly calculated.

Finally, two additional features we employed to enrich the
information that is available to the DL model. The first one is
a binary variable that is used to indicate whether the current
day is a public holiday. The other feature was used to indicate
whether the time slot belongs to a period of a pandemic or
not. We included this feature since we observed that during
the recent COVID-19 pandemic it was observed that traffic
was affected. Even though we used this feature to only mark
time slots that belong to COVID-19 pandemic, we suspect that
this feature might be also useful during subsequent flu seasons
due to increased public awareness. 

All features were normalized using standardization, i.e., subtracting their mean and dividing by the standard deviation. The training set statistics were used for the normalization process. Also, the proposed prediction system operates on one-hour time slots aiming to predict the parking violation rate for each time slot and sector. The target parking violation rate is calculated as:
\begin{equation}
p = \frac{N_s}{N_c}, \tag{5}
\end{equation}
where \( p \) denotes the target violation rate to be predicted, \( N_s \) is the number of illegally parked cars and \( N_c \) is the capacity of the specific sector at a specific time slot (again we omit the sector and time slot indices to avoid cluttering the used notation). The number of illegally parked cars is calculated from police scan data. During such scans, a police officer scans cars’ license plates that are parked in a sector. The scanner then checks if the parking fee has been paid (or if the car can legally park there without paying a fee) for each parked car. Then, this information is used to calculate the number of illegally parked cars within each sector and time slot. Then, each scan can be assigned to the time slot that it is nearest to the center of each time slot.

To this end, we applied a Gaussian-based smoothing scheme. After calculating the violation rate for each scanning session, we distributed the observed violation rate to nearby time slots using the following equation:
\begin{equation}
y = \frac{1}{|\mathcal{S}|} \sum_{s \in \mathcal{S}} \exp\left(-\frac{d_s}{\sigma}\right)p_s, \tag{6}
\end{equation}
where \( s \) denotes a scanning session, \( \mathcal{S} \) consists of scanning sessions overlapping or being nearby (by a difference of one time slot) to the current time slot, \( d_s \) is the distance in minutes of scanning session \( s \) from the current time slot’s center, \( p_s \) is the violation rate calculated for scanning session \( s \), and \( |\mathcal{S}| \)

\begin{figure}[H]
    \centering
    \includegraphics[width=0.5\linewidth]{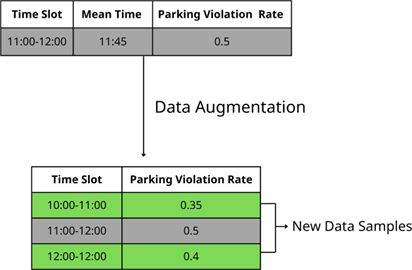}
    \caption{The proposed data augmentation and smoothing method allows for assigning observed violation rates into nearby time slots, reducing the impact of sparse annotations and missing data scans.}
    \label{fig:2}
\end{figure}
Denotes the cardinality of set \( \mathcal{S} \), i.e., the number of scanning sessions that can be assigned to the current time slot. The parameter \( \sigma \) controls the distribution process, i.e., a larger value distributes the violations to the nearby time slots, while smaller values allow for more steep changes in the parking violation rate between slots. A value of \( d = 210 \) minutes was used for the models developed in this paper. For example, if we observe a rate of thirty five percentage (35\%) of parking violations in a sector at twelve and fifty (12:50), i.e., in “12:00–13:00” time slot, then we can also assume that this rate should be also partially assigned to the next time slot of “13:00–14:00”, as shown in Fig. 2. As described before, this allows for filling potential gaps in the data, reducing the impact of sparse annotations and missing scans.

In this work, we employed a residual DL architecture for predicting the parking violation rate in different sectors. The employed architecture is composed of six hidden layers (each one with 512, 256, 128, 64, 128 and 32 neurons respectively) and one output layer. The ReLU activation function was used for the hidden layer~\cite{ref10}, while a residual link was used between the third and fifth hidden layers~\cite{ref11}. Using this link allows for improving the prediction accuracy over architectures that do not use residual connections. Also, further increasing the size of the employed architecture is not expected to lead to further accuracy improvements. The last layer employs a sigmoid activation function, since it is used to predict the parking violation rate, which is always bounded between 0 and 1. To avoid pushing the sigmoid function to its extremes, we re-normalized targets between 0.1 and 0.9. The Adamax optimization algorithm was used for training the model~\cite{ref12}. The learning rate was initialized at 0.001 and an exponential learning rate decay strategy with a decay equal to 0.25 was used for performing learning rate scheduling~\cite{ref13}.

\section{Experiments}
For all the conducted experiments, we used data collected from Thessaloniki’s on-street public parking system called THESi.
THESi consists of about 4700 parking slots in the

\begin{figure}[H]
    \centering
    \includegraphics[width=0.5\linewidth]{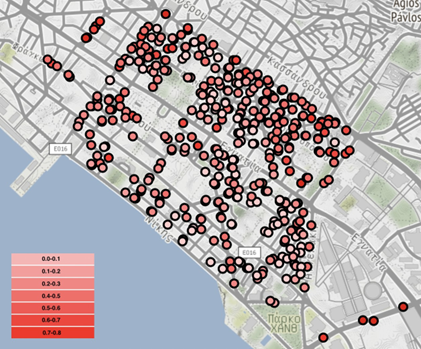}
    \caption{Spatial distribution of parking violations. Darker values indicate higher average parking violation rate per sector}
    \label{fig:3}
\end{figure}
City’s centre, which are distributed into 396 separate sectors. The aim of the proposed method is to predict the parking violation rate for each of these sectors. We used historical data that consist of scans conducted by the city’s traffic police, as well as weather data from OpenWeather. A total of 3.8 million scans that were captured through 300,000 checks were used in the conducted evaluation. Furthermore, we manually defined 19 PoIs used for encoding the distances for each sector, as described in Section II. Also, we used 80\% of the data to train the model, while the remaining 20\% was used to test the model. The mean absolute error (MAE) was used as the evaluation metric.

We provide a brief data analysis before using the collected data for predicting parking violations. Parking violation rate ranges between zero (0) and one (1), while the average parking violation rate is 0.41. Given that an average sector has eleven parking slots, this parking violation rate means that three to four (3-4) parking slots are expected to be illegally occupied. We also examined the spatial distribution of parking violations in Fig. 3, where darker colors indicate a higher parking violation rate. The higher mean parking violation of a sector is 0.78, while the lower is 0.1. We observe a higher parking violation rate in the upper region of the city. Another observation is that parking violations are higher in remote sectors. This can be probably explained since drivers might believe that these sectors are not as frequently scanned for illegal parking.

The evaluation results are reported in Table I. The baseline model without any smoothing achieves a MAE of 0.175, which demonstrates that we can accurately predict the parking violation rate using the proposed method, since the MAE of a baseline average value predictor (using the average violation rate as the prediction) is 0.251. Furthermore, using the proposed data augmentation and smoothing method for the train set allowed for further reducing the error to 0.169 on the original test set, while the error dropped to 0.146 when the same smoothing methodology was applied to the whole dataset. These results demonstrate the benefit of employing the proposed data augmentation and smoothing approach to fill
\begin{figure}[H]
    \centering
    \includegraphics[width=0.5\linewidth]{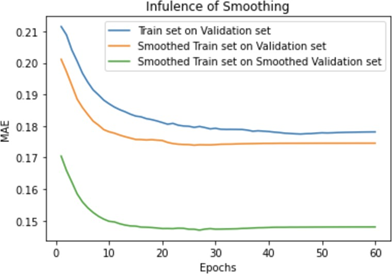}
    \caption{Validation error during training when using two different setups and evaluation sets.}
    \label{fig:4}
\end{figure}
\begin{table}[h]
\centering
\caption{Parking Violation Rate Prediction Evaluation}
\begin{tabular}{lcc|c}
\toprule
\textbf{Method} & \textbf{Raw Test Set (MAE)} & \multicolumn{1}{c|}{} & \textbf{Smoothed Test Set (MAE)} \\
\midrule
Without Smoothing & 0.175 & & 0.173 \\
With Smoothing    & \textbf{0.169} & & 0.146 \\
\bottomrule
\end{tabular}
\end{table}

missing data and reduce data discontinuities.

We also conducted a 4-fold cross validation experiments, where we monitored the validation MAE during the training epochs, to select the most appropriate number of training epochs for each model. The results are provided in Fig. 4. Note that using the smoothed training set leads to significant improvements over directly using the raw training set. We also plotted the MAE for the smoothed validation set, where again we observe that the training process converges smoothly, while achieving significantly better MAE, confirming the results reported in Table I.

\section{Conclusions}
In this work, we presented and evaluated a DL-based model for on-street parking violation prediction. Several challenges were faced during the development of such a system, ranging from data sparsity challenges to encoding data in the most appropriate format for the employed DL model. To address these issues, we developed a novel approach that assigns the existing ground truth annotation to spatial sectors and time slots, leading to significant performance improvements. Other challenges include imprinting input data, such as the geographical location of each sector, time, and other information, such as pandemic measures (i.e., due to COVID-19) in an efficient manner. We evaluated the proposed method using novel data from the THESi system implemented in Thessaloniki, Greece, while we also presented various aspects of the employed data to provide more insight into this especially challenging problem.

There are several interesting future research directions. First, graph neural networks can be used to better model spatial relations between sectors~\cite{ref14}. Furthermore, traffic on city streets can affect the rate of parking violations, hinting that this could be an additional feature that can be included to further enhance the performance of the presented model. Furthermore, weather and mobility information can be a valuable predictor, although such data typically is not available for all sectors/roads, highlighting the need for systems that can work even under the presence of sparse information.

\end{document}